\documentclass[10pt,twocolumn,letterpaper]{article}

\usepackage{wacv}
\usepackage{times}
\usepackage{epsfig}
\usepackage{graphicx}
\usepackage{amsmath}
\usepackage{amsthm}
\usepackage{hhline}
\usepackage{amssymb}
\usepackage{multirow}
\usepackage{bbm}
\usepackage{float}
\restylefloat{table}

\wacvfinalcopy 

\def\wacvPaperID{****} 

\ifwacvfinal\fi
\begin{document}
\title{Frustum VoxNet for 3D object detection from RGB-D or Depth images}
\author{Xiaoke Shen\\
The Graduate Center, CUNY\\
New York City, USA\\
{\tt\small xshen@gradcenter.cuny.edu}
\and
Ioannis Stamos\\
Hunter College \& The Graduate Center, CUNY\\
New York City, USA\\
{\tt\small istamos@hunter.cuny.edu}
}
\clearpage\maketitle
\thispagestyle{empty}

\ifwacvfinal\pagestyle{empty}\fi
\begin{abstract}
Recently, there have been a plethora of classification and detection systems from RGB as well as 3D images.
In this work, we describe a new 3D object detection system from an RGB-D or depth-only point cloud. 
Our system first detects objects in 2D (either RGB, or pseudo-RGB constructed from depth). The next step is to detect 3D objects within the 3D frustums 
these 2D detections define. This is achieved by voxelizing parts of the frustums (since frustums can be really large), instead of using the whole frustums as done in earlier work.
The main novelty of our system has to do with determining which parts (3D proposals) of the frustums to voxelize, thus allowing us to provide high resolution representations around the objects of interest. It  also allows our system to have reduced memory requirements. These 3D proposals are fed to an efficient ResNet-based 3D Fully Convolutional Network (FCN). 
Our 3D detection system is  fast, and can be integrated into a robotics platform. With respect to systems that do not perform voxelization (such as PointNet), our methods can operate without the requirement of subsampling of the datasets. We have also introduced a pipelining approach that further improves the efficiency of our system.
 Results on SUN RGB-D dataset show that our system, which is based on a small network, can process 20 frames per second with comparable detection results to the state-of-the-art \cite{DBLP:journals/corr/abs-1711-08488}, achieving a $2\times$ speedup.
\end{abstract}
\section{Introduction}
\begin{figure}[t]
\begin{center}
   \includegraphics[width=1.0\linewidth]{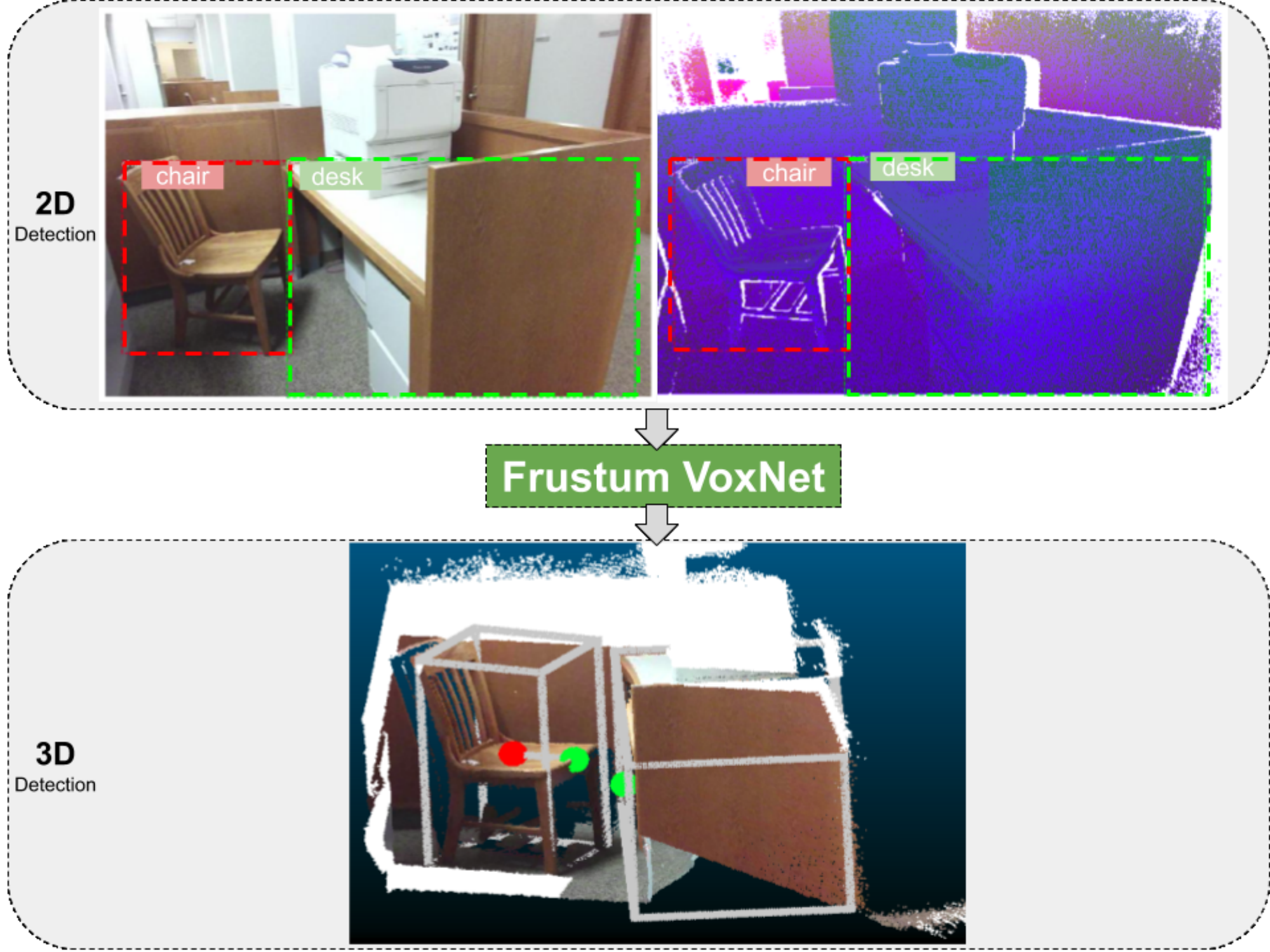}
\end{center}
	\caption{Overview of the whole system. Upper left: RGB image and detected 2D bounding boxes. Upper right: DHS (\textbf{D}epth \textbf{H}eight and \textbf{S}igned angle) image, and detected 2D bounding boxes. A DHS image is a pseudo-RGB image generated by a depth image (see text). 
	Bottom: The final 3D detected objects from the associated 3D range image. The 3D detection not only provides an amodal bounding box but also an orientation. The red point is the center of the bounding box and the green one is the front center. 
	The detected 2D bounding boxes from either and RGB or DHS image, generate 3D frustums (which are prisms having as apex the sensor location and extend through the 2D bounding boxes to the 3D space). They are then fed to our Frustum VoxNet network, which produces the 3D detections. }
\label{fig:whole}
\end{figure}
 Classification and object detection are significant problems in the fields of computer vision and robotics.
 2D object detection systems from RGB images have been significantly improved in recent years due to the emergence of deep neural networks and large labeled image datasets. For applications related to robotics though, such as autonomous navigation, grasping, etc., a 2D object detection system is not adequate. Thus 3D object detection systems have been developed, with input coming from RGB-D or depth-only sensors. In this paper we describe a new 3D object detection system that incorporates mature 2D object detection methods as a first step. The 2D detector can run on an input RGB image, or pseuso-RGB image generated from a 3D point cloud. That 2D detection generates a 3D frustum (defined by the sensor and the 2D detected bounding box) where a search for a 3D object is performed. Our main contribution is the 3D object detection within such as frustum. Our method involves 3D voxelization, not of the whole frustum, but of a learned part of it. That allows for a higher resolution voxelization, lower memory requirements, and a more efficient detection.
  
  Figure \ref{fig:whole} illustrates the overview of our system. In the upper left we see a 2D RGB image, along with the 2D detected bounded boxes (a chair and a desk). On the upper right we see a 2D pseudo-RGB image that was generated from the associated 3D range image (see \cite{7785116}), along with similarly detected 2D bounded boxes. We call this pseudo-RGB image a {\bf DHS} image, where D stands for \textbf{D}epth, H for \textbf{H}eight, and S for \textbf{S}igned angle. The depth is a normalized distance of the associated 3D point, height is a normalized height of the 3D point, and the signed angle is a normalized approximation of the normal at the 3D point (see \cite{7785116}). We can apply traditional 2D detectors on this pseudo-RGB image, making our method applicable even when no RGB information is available. 3D frustums are then extracted from these 2D detections. A 3D frustum is a prism having as apex the sensor location and extending through the 2D bounding boxes into the 3D space. Learned parts of the 3D frustum are being voxelized. These voxelizations are fed to Frustum VoxNet, which is a 3D Fully Convolutional Neural Network (FCN).
    
The key contributions of our work can be summarized as follows:
\begin{itemize}
\item We demonstrate the power of using a 3D FCN approach based on volumetric data to achieve accurate 3D detection results efficiently.
\item We provide a novel method for learning the parts of 3D space to voxelize. This allow us to provide high resolution representations around the objects of interest. It also allows our system to have reduced memory requirements and leads to its efficiency.
\item Compared to systems that do not perform voxelization (such as \cite{DBLP:journals/corr/QiSMG16, DBLP:journals/corr/abs-1711-08488}), our methods can operate without the requirement of subsampling the datasets. Also, our approach is more efficient and can be used in robotics applications.
\item Compared to systems that do voxelize (such as \cite{Zhou-2017-VoxNet}), our system does not voxelize the whole space, and thus allows a higher-resolution object representation.
\item We compare the 3D detection performance of using different input channels (RGB or DHS).
\item We provide a more efficient variation of our method that involves pipelining, geared to robotics applications.
\item The parameters of our network are much smaller than leading methods. That results in faster inference time.
\end{itemize}

We start by reviewing related work and then proceed with the description of our 3D detection system along with our experimental results. Since our final goal is indoor robotic navigation, our current system has been evaluated based on an indoor SUN-RGBD dataset\cite{Song_2015_CVPR}. 

 \section{Related Work}
\textbf{2D methods}  RGB-based approaches can be summarized as two-stage frameworks (proposal and detection stages) and one-stage frameworks (proposal and detection in parallel). Generally speaking, two-stage methods such as R-CNN \cite{DBLP:journals/corr/GirshickDDM13}, Fast RCNN \cite{DBLP:conf/iccv/Girshick15}, Faster RCNN \cite{DBLP:conf/nips/RenHGS15}, FPN \cite{DBLP:journals/corr/LinDGHHB16} and mask R-CNN \cite{DBLP:journals/corr/HeGDG17} can achieve a better detection performance while one-stage systems such as YOLO\cite{DBLP:journals/corr/RedmonDGF15}, YOLO9000\cite{DBLP:journals/corr/RedmonF16} and RetinaNet \cite{DBLP:journals/corr/abs-1708-02002} are faster at the cost of reduced accuracy. For deep learning based systems, as the size of network is increased, larger datasets are required. Labeled datasets such as PASCAL VOC dataset \cite{pascal-voc-2012} and COCO (Common Objects in Context) \cite{DBLP:journals/corr/LinMBHPRDZ14} have played important roles in the continuous improvement of 2D detection systems.

\textbf{3D methods}  Compared with detection based on 2D images, the detection based on 3D data is more challenging due to several reasons \cite{shen2019}: 1) Data representation itself is more complicated. 3D images can be represented by point clouds, meshes, or volumes, while 2D images have pixel grid representations. 2) Due to the extra dimension, there are increased computation and memory resource requirements. 3) 3D data is generally sparser and of lower resolution compared with the dense 2D images, making 3D objects more difficult to identify.  Finally, 4) large sized labeled datasets, which are extremely important for supervised based algorithms, are still inferior compared with well-built 2D datasets. Below we summarize the basic approaches.

\textbf{Project 3D data to 2D and then employ 2D methods} There are different ways to project 3D data to 2D features. HHA was proposed in \cite{DBLP:journals/corr/GuptaGAM14} where the depth image is encoded with three channels: Horizontal disparity, Height above ground, and the Angle of each pixel’s local surface normal with gravity direction. The signed angle feature described in\cite{6375012} measures the elevation of the vector formed by two consecutive points and indicates the convexity or concavity of three consecutive points. Input features converted from depth images of normalized depth(D), normalized relative height(H), angle with up-axis(A), signed angle(S), and missing mask(M) were used in \cite{7785116}. We are using DHS in this work to project 3D depth image to 2D since as shown in \cite{7785116} adding more channels did not affect classification accuracy significantly. Keeping the number of total channels to three, allow us to use networks with pre-trained weights for starting our training.

\textbf{2D-Driven 3D Object Detection from RGB-D Data}
Our proposed framework is mainly inspired by 2D-driven 3D object detection approaches as in \cite{Lahoud_2017_ICCV, DBLP:journals/corr/abs-1711-08488}. First a 2D detector is used to generate 2D detections. The differences of our work with \cite{Lahoud_2017_ICCV} are: 1) the 2D detector in \cite{Lahoud_2017_ICCV} is only based on RGB images and our proposed system explores both RGB-D and Depth only data. 2) 3D detection in \cite{Lahoud_2017_ICCV} uses a MLP regressor to regress the object boundaries based on histograms of points along $x$, $y$, and $z$ directions. Converting raw point clouds to histograms results in a loss of information. The main differences of our system to Frustum PointNets \cite{DBLP:journals/corr/abs-1711-08488} are the following: 1) in the 2D detection part, Frustum PointNets is based on RGB inputs, while our system can support both RGB-D and depth-only sensing. 2) in the 3D detection part, our system is using voxelized data, while Frustum PointNets is consuming raw point clouds via PointNet \cite{DBLP:journals/corr/QiSMG16}. PointNet uses a fully connected neural network and max pooling, so it cannot support  convolution/deconvolution operations well. We believe 3D convolution/deconvolution can play important roles in both 3D semantic segmentation and object detection. 3) PointNet's computation complexity is increased if more points are available as the framework's input is $N \times K$ where $N$ is the number of points and $K$ is the number of channels. 4) Random sampling is required in PointNet, but is not needed in our voxelization approach.

 A recent method \cite{qi2019deep} that is based on PointNet and Hough Voting, achieves improved detection results without the use of RGB images. Our method is still more efficient in inference time, and thus more appropriate for robotics application. Also, our approach does not need to subsample the 3D point cloud as required by \cite{qi2019deep}.

\textbf{3D CNNs} VoxelNet \cite{Zhou-2017-VoxNet} uses 3D LiDAR data to detect 3D objects based on the KITTI outdoor dataset, and utilizes bird's eye view (BEV) features (such as MV3D \cite{DBLP:journals/corr/ChenMWLX16} and AVOD \cite{2017arXiv171202294K})). The use of BEV is not helpful in indoor applications. Also, the use of the whole range image for voxelization lowers the resolution (and therefore the scale) of the objects of interest.
    Early influential 3D detection systems used two-stage approaches. The first stage generates proposals, while the second stage performs 3D detection. DeepSliding Shape\cite{DBLP:journals/corr/SongX15} detects 3D objects based on the SUNRGB-D dataset and it uses directional Truncated Signed Distance Function (TSDF) to encode 3D shapes. The 3D space is divided into 3D voxels and the value in each voxel is defined to be the shortest distance between the voxel center and the surface from the input depth map. 
    A fully convolutional 3D network extracts 3D proposals at two scales corresponding to large size objects and small size objects. For the final 3D detection, this method fuses the 3D voxel data and RGB image data by using 3D and 2D CNNs. Our approach, one the other hand, first focuses on the frustum to voxelize, and then selects the part to be voxelized based on training. That allows us to achieve higher resolution around the objects of interest.
    
We refer readers to \cite{shen2019} for latest, comprehensive comparisons of different 3D detection systems.
\section{Dataset}
We are focusing on the indoor SUN RGBD dataset\cite{Song_2015_CVPR}. SUN RGBD dataset splits the data into a training set which contains 5285 images and a testing set which contains 5050 images. For the training set, it further splits into a training only, which contains 2666 images and a validation set, which contains 2619 images. Similar to \cite{DBLP:journals/corr/SongX15,Lahoud_2017_ICCV}, we are training our model based on the training only set and evaluate our system based on the validation set. We call the only training dataset as train2666 in the future description.\\
%
\section{Frustum VoxNet System Overview}
First, 2D detections on RGB or DHS image generate 2D bounding boxes of objects. The 2D detections generate 3D frustums (defined by the sensor and the 2D detected bounding box) where a search for a 3D object is performed. For each such frustum we know the class of the object to be detected by the 2D detection. Our system accurately localizes the amodal 3D bounding box and the orientation of the detected 3D object. To achieve this, we perform 3D voxelization, not of the whole frustum, but a learned part of it. That allows for a higher resolution voxelization, lower memory requirements, and a more efficient detection. We explain first how we decide which part of the frustum to use.

\subsection{Frustum Voxelization}
Given a 3D frustum (defined as a 3D prism from the sensor and the 2D detected bounding box into the 3D space), our goal is to voxelize only a part of it. We define that part as axis-aligned 3D bounding boxes enclosed in the frustum. We call that bounding box a {\bf 3D Cropped Box (3DCB} for short).
Given a specific object class (for instance a table), an ideal 3DCB will be big enough to contain all the 3D points belonging to the object, but also small enough to achieve high resolution voxelization. In order to quantify the ability of a given 3DCB to tightly contain a given 3D object, we define the metric \textit{3D Intersection over Itself (IoI)}. Suppose the object of interest lies in a bounding box 3DBBOX. Then the IoI of the 3DBBOX wrt to a given 3DCB is defined as the volume of intersection of the 3D bounding box with the 3DCB over the volume of the 3D bounding box itself. Therefore an IoI of 1.0 means that the 3DCB is perfectly enclosing the object in 3DBBOX, while as this number tends to 0.0 more and more empty space is included in the 3DCB.

The formula for 3D IoI is:\\
$$IoI^{3D}=\frac{volume^{3DBBOX} \cap volume^{3DCB}}{volume^{3DBBOX}}$$
From the definition, it is trivial to show that:\\
$$IoI^{3D}=IoI^{XY}*IoI^{Z}$$
where $IoI^{XY}$ is the IoI in the $XY$ plane and $IoI^{Z}$ is the IoI along the $Z$ axis.\\
$$IoI^{XY}=\frac{area^{3DBBOX^{XY}}\cap area^{3DCB^{XY}}}{area^{3DBBOX^{XY}}}$$
$$IoI^{Z}=\frac{length^{3DBBOX^{Z}}\cap length^{3DCB^{Z}}}{length^{3DBBOX^{Z}}}$$
$3DBBOX^{XY}$ and ${3DCB^{XY}}$ are 2D projections of 3D bounding box and 3DCB onto the $XY$ plane. $3DBBOX^{Z}$ and ${3DCB^{Z}}$ are 1D projections of 3D bounding box and 3DCB onto the $Z$ axis.\\
\begin{figure}[t]
\begin{center}
   \includegraphics[width=0.5\linewidth]{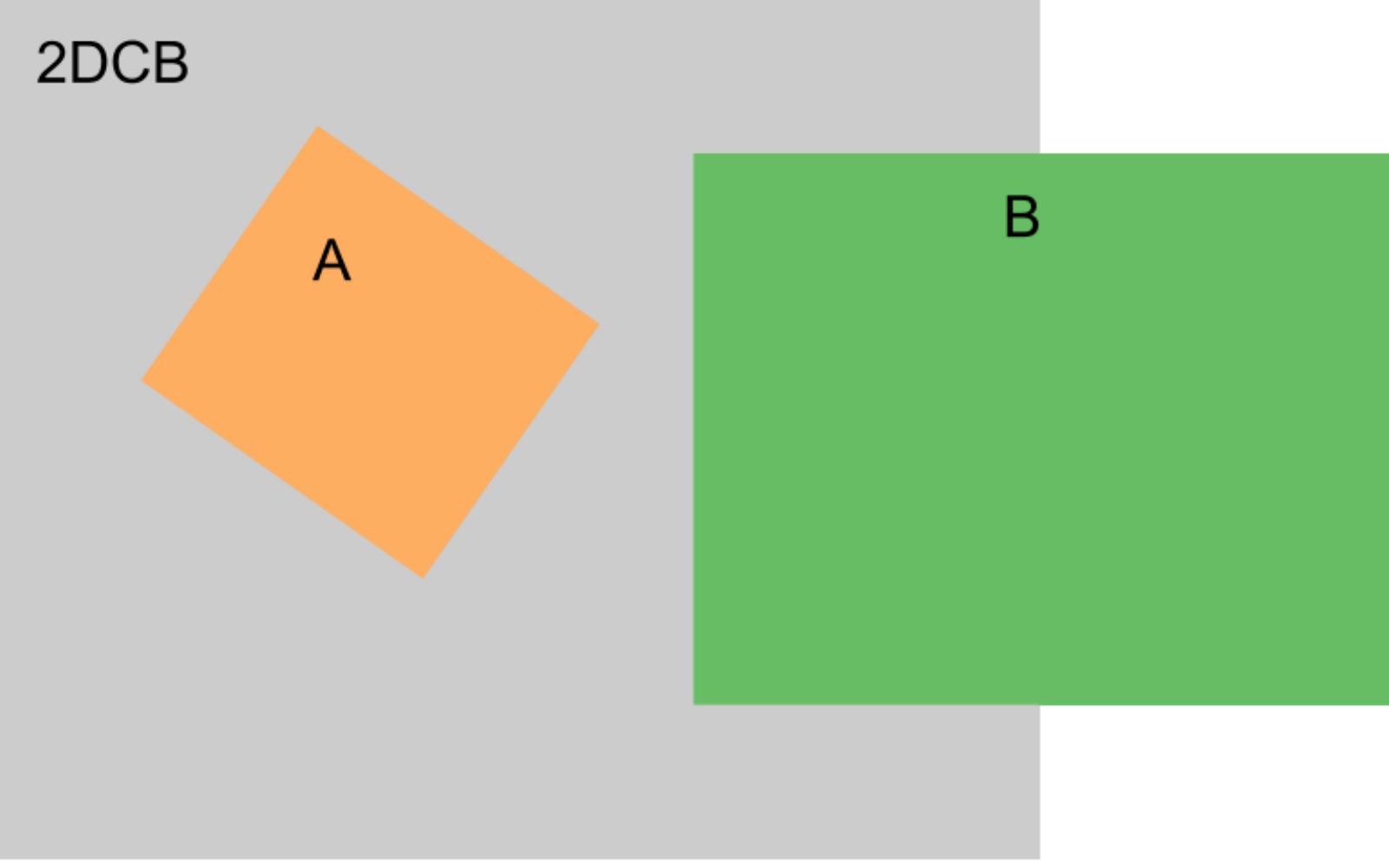}
\end{center}
	\caption{An example of 2DCB with two objects box A and box B. All these boxes are square. A has length 1, B has length 2 and 2DCB has length 3. Half of B is overlapped with 2DCB}
\label{ioi_iou}
\end{figure}
We use this metric to choose the optimal 3DCB size.
A 2D example in Figure \ref{ioi_iou} is used to show the difference between IoI and IoU (Intersection over Union). From this example, box A is totally contained in 2DCB($XY$ plane projection of a 3DCB) while only half of box B is covered by 2DCB. If we use 2D IoU, we will get 0.11 for box A with 2DCB and 0.18 for box B with 2DCB.
\begin{figure}[t]
\begin{center}
   \includegraphics[width=1.0\linewidth]{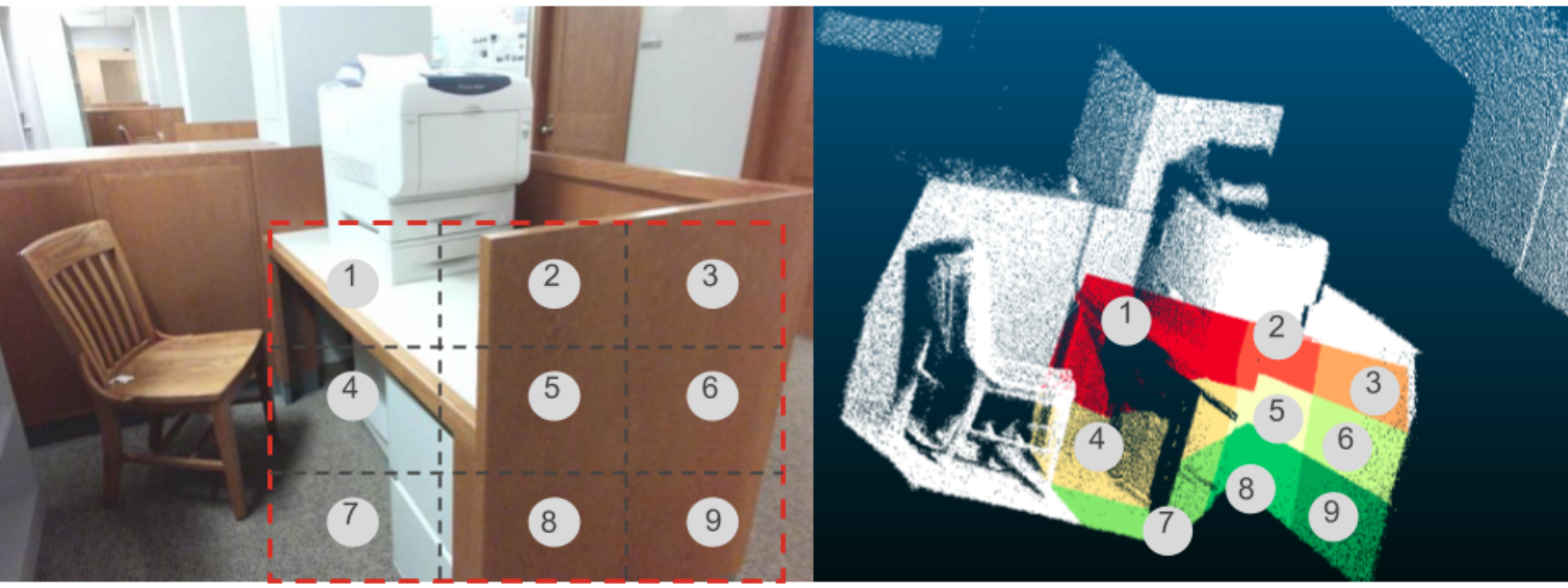}
\end{center}
        \caption{An example of equally subdividing a whole frustum into $3 \times 3$ subfrustums (best viewed in color). In this example, the object is a desk. The upper one shows the 2D bounding box of desk is equally divided into 9 small boxes. From each small box, a subfrustum is generated as shown in the bottom image.}
\label{subfrustum}
\end{figure}

\textbf{Generating 3DCBs using an IoI metric}
During training, given a ground truth 2D bounding box of an object of a given class (for example table) and given the ground truth 3D bounding box of the same object, we would like to calculate the optimal 3DCB box. The 3DCB is represented by its center, and width, depth, and height. We are adding the constraint that width and depth are the same. This makes sure that the object can freely rotate within the 3DCB along the vertical axis. 
We proceed by equally dividing the 2D bounding box along the \textbf{R}ow and \textbf{C}olumn into $FR \times FC$ 2D boxes. Then we have $FR \times FC$ sub\textbf{F}frustums. We will generate $FR \times FC$ candidate centers of 3DCBs in that case. The center of each 3DCB is the centroid of the respective frustum. One example of $3 \times 3$ subfrustums of a desk is shown in Figure \ref{subfrustum}. If we set $FR=FC=1$, then there is only one 3D frustum to consider (and therefore one 3DCB center). Our goal is to calculate the optimal sizes of respective 3DCBs for each object category.

A ground truth 3D bounding box will be recalled (i.e. enclosed into the 3DCB) if the \textit{3D IoI} of this box is greater than a threshold. Formally, we define this recall as $recall^{volume}$:\\
$$recall^{volume} = \frac{|3DCB^{positive}|}{|3DCB|}$$
where $|3DCB^{positive}|$ is the cardinality of positive 3DCBs and $|3DCB|$ is the cardinality of all 3DCBs. A 3DCB is positive when $IoI^{3D} = IoI^{XY} * IoI^{Z} \geq threshold$.
To make the parameter setting simple, we are exploring the recall of $XY$ plane and $Z$ axis separately. Similar to $recall^{volume}$, $recall^{XY}$ and $recall^{Z}$ are defined as: $recall^{XY} = \frac{|3DCB^{positive}_{XY}|}{|3DCB|}$, $recall^{Z} = \frac{|3DCB^{positive}_{Z}|}{|3DCB|}$, where $|3DCB^{positive}_{XY}|$ is the cardinality of positive 3DCBs in XY plane, $|3DCB^{positive}_{Z}|$ is the cardinality of positive 3DCBs in Z axis and $|3DCB|$ is the cardinality of all 3DCBs. A 3DCB is positive in $XY$ plane when $IoI^{XY} \geq threshold_{XY}$ and a 3DCB is positive in $Z$ axis when $IoI^{Z} \geq threshold_{Z}$.\\
Although, we can NOT naively have $recall^{volume} =recall^{XY} *recall^{Z}$, we have a nice inequality to guarantee a lower bound of $recall^{volume}$:\begin{equation}  \label{eq:1}
 recall^{volume} \geq \max(0, \quad recall^{XY} +recall^{Z} -1)
\end{equation}
\begin{figure}[t]
\begin{center}
   \includegraphics[width=0.8\linewidth]{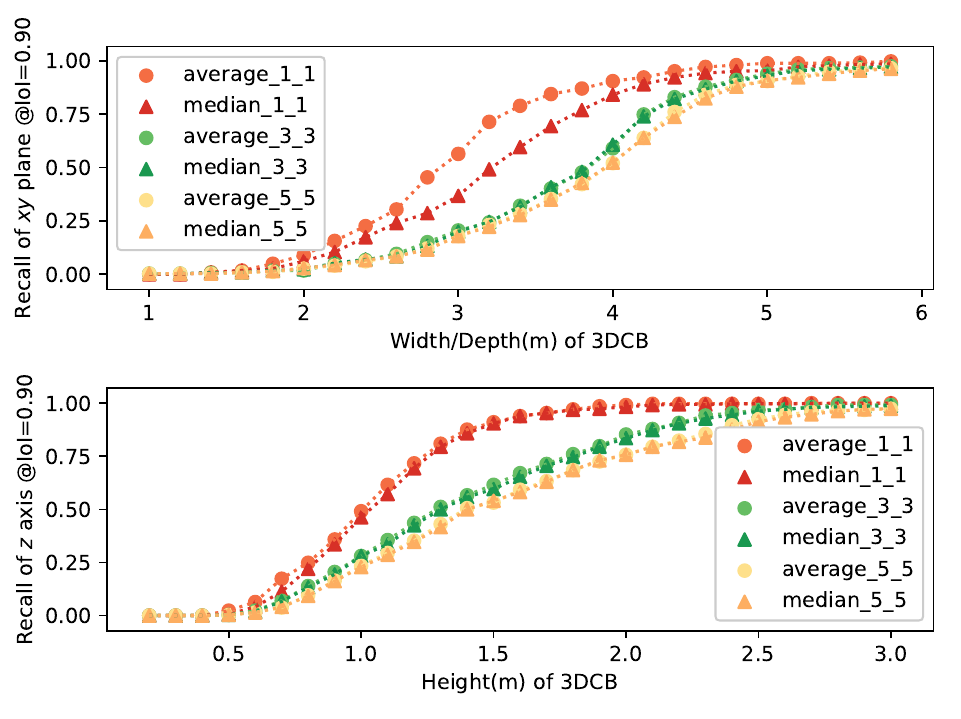}
\end{center}
        \caption{$IoI^{XY}$ and $IoI^{Z}$ with the widths/depths and heights. 3DCB are generated from average/median center based on $FR \times FC$ subfrustums with different widths/depths and heights. In this plot, \textit{average/median$\_$m$\_$n} corresponds to recall based on average/median center in $m \times n$ subfrustums.}
        \label{xyz_recall}
\end{figure}
The proof of this inequality is given in the appendix.

Both of $threshold_{XY}$ and $threshold_{Z}$ are set as $0.90$. We are generating both the average center and median center from subfrustums and pick up the best one from these $FR \times FC$ candidates to calculate the recall. The average recall based on different setups of width/depth and height are shown in Figure \ref{xyz_recall}. From the results, we can observe: 1) the performance of the average center based 3DCB is better especially when $1 \times 1$ subfrustums are used compared with the median center. The reason for this might be the range of indoor depth sensor is limited and outliers will not have too much influence to the results. 2) The 3DCB generated from $1 \times 1$ is better than $3 \times 3$ and $5 \times 5$ ones. Based on these observations, we are choosing both $1 \times 1$ and $3 \times 3$ during training to generate more samples and make the training robust to the inaccurate bounding box predictions. During inference, $1 \times 1$ subfrustum based 3DCB is used to speed up and get better performance.\\ 

\textbf{Double Frustum Method} 
\label{sec_double} To increase the accuracy of the center calculations, we developed a double Frustum framework. We use a smaller 2D bounding box to generate a smaller frustum for the calculation of the 3DCB center. The estimated center should now be more accurate since it will concentrate on the central part of the object and thus will avoid the use of other background objects. A 3DCB is then selected from a larger frustum in order to contain background context points and possible false negative points. The larger frustum is generated from a larger 2D bounding box. During training, we generate large frustum by randomly increasing the 2D bounding box width and height by $0\%$ to $15\%$ independently. For the small frustum, we randomly decrease the 2D bounding box width and height by $0\%$ to $10\%$ independently. During inference, the large frustum is generated by increasing the 2D bounding box width and height by $5\%$. Original 2D detection bounding boxes are used to calculate the 3DCB center. 

\textbf{Multiple Scale Networks}
In \cite{DBLP:journals/corr/SongX15}, two scales network were used for different categories concerning the 3D physical size. We are using 4 scales networks to voxelize the 3D objects corresponding to the average physical size of average height, maximum of average width and depth. The mapping of 3D object categories to different scales is shown in Table \ref{mul_scale_network}. 
\begin{table}[t]
\begin{center}
		\resizebox{\linewidth}{!}{
\begin{tabular}{|c|c|c|}
\hline
	 & Short ($h \leq 0.55$)& Tall ($h > 0.55$)\\
\hline
\hline
	\shortstack{Small ($\max(w,h) \leq 0.3$)}& \shortstack{toilet} &N/A\\
\hline
	\shortstack{Medium\\($0.3 < \max(w,h) \leq 0.55$)}& \shortstack{chair, nightstand, sofa chair,\\garbage bin,bathtub} &\shortstack{bookshef}\\
\hline
	 \shortstack{Large\\($\max(w,h) > 0.55$)}& \shortstack{table, desk, \\sofa, bed, dresser}& N/A\\
\hline
\end{tabular}}
\end{center}
\caption {Objects are classified into 4 categories based on there average physical size. Voxelization is processed based on each category.}
\label{mul_scale_network}
\end{table}
\begin{figure}[t]
\begin{center}
   \includegraphics[width=0.9\linewidth]{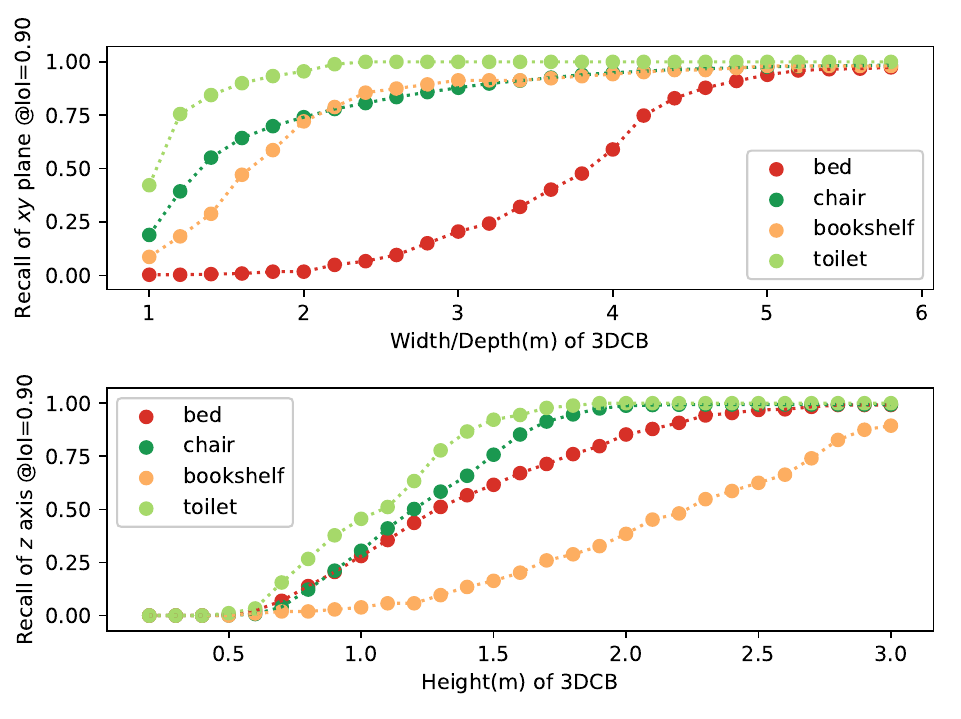}
\end{center}
	\caption{$XY$ plane recall and $Z$ axis recall for bed, chair bookshelf and toilet with the widths/depths and heights based on train2666 dataset.}
	\label{xyz_recall_more}
\end{figure}
We are calculating the $recall_{XY}$ and $recall_{Z}$ for different objects with the different setups for width/depth and heights. The curves of $recall_{XY}$ with width/depth and $recall_{Z}$ with height are plotted for four classes based on $3 \times 3$ subfrustums (sofa is from large short scale, chair is from medium short scale toilet is from small short scale and bookshelf is from median tall scale) are shown in Figure \ref{xyz_recall_more}. From these curves, we can find out that medium tall scale category needs greater height and both the large short and medium short categories need more width/depth. We are selecting the minimum width/depth and height which can guarantee all objects within that scale network can meet the requirements of $recall^{XY} \geq 0.90$ and $recall^{Z} \geq 0.95$. This is based on $3 \times 3$ subfrustums. From the equation \ref{eq:1}, we can have the lower bound of the $recall_{volume}$ of $0.85$. Although $0.85$ is not high enough, when based on $1 \times 1$ subfrustums, the lower bound of the $recall_{volume}$ can achieve $0.94$ as $recall^{XY} \geq 0.95$ and $recall^{Z} \geq 0.99$ for 3DCBs generating from $1 \times 1$. Since we are using both 3DCBs from $1 \times 1$ and  $3 \times 3$ subfrustums, the recall is good enough to support the training.\\
\begin{table}[t]
\begin{center}
		\resizebox{\linewidth}{!}{
\begin{tabular}{|c|c|c|c|c|}
\hline
	Method& Network& \shortstack{3DCB \\physical size\\(m)}&\shortstack{3DCB \\Shape}&\shortstack{Resolution\\(cm)}\\
\hline
\hline
	\multirow{3}{*}{\shortstack{DSS \\ \cite{DBLP:journals/corr/SongX15}}}&RPN &$2.5\times2.5\times2.5$&$208\times208 \times 100$&$5.2\times6.0\times 2.5$\\
	\hhline{~----}
	&\shortstack{Detection\\(bed)} &$6.7\times6.7\times3.2$&$30\times30\times 30$&$2.0\times2.0\times 0.95$\\
	\hhline{~----}
	&\shortstack{Detection\\(trash can)} &$1.0\times1.0\times1.2$&$30\times30\times 30$&$0.3\times0.3\times 0.5$\\
\hline\hline
	\multirow{4}{*}{\shortstack{Ours}}&small short&$ 1.6\times 1.6\times 1.5$&$198\times198\times 102$&$ 0.8\times0.8 \times1.5 $\\
	&medium short&$3.2\times3.2\times1.7$&$198\times198\times 102$&$1.6\times1.6\times 1.7$\\
	\hhline{~----}
	&large short&$4.8\times4.8\times2.2$&$198\times198\times 102$&$2.4\times2.4\times 2.2$\\
	\hhline{~----}
	&medium tall&$2.8\times2.8\times3.0$&$134\times134\times 134$&$2.1\times2.1\times 2.2$\\
\hline
\end{tabular}}
\end{center}
\caption {Resolution and shape comparison between DeepSliding Shape\cite{DBLP:journals/corr/SongX15} and ours. Anchors of the bed and trash can from\cite{DBLP:journals/corr/SongX15} are used as examples of proposal's physical size to make the comparison with ours.}
\label{big_vox_compare}
\end{table}
The physical sizes(width/depth/height) of 4 scale networks are shown in Table \ref{big_vox_compare} based on the principles described above. 3DCB are further voxalized(counting the number of cloud points within each voxel) into a 3D tensor with the shape of $W \times D \times H$. The $W \times D \times H$ for each scale network are selected to make it having a better resolution as compared with \cite{DBLP:journals/corr/SongX15}. The comparison of physical size, resolution, tensor shape of the RPN and detection networks of \cite{DBLP:journals/corr/SongX15} and ours are also shown in Table \ref{big_vox_compare}.\\
\subsection{3D Object Detection}
\textbf{3D Bounding Box Encoding}
Similar to \cite{DBLP:journals/corr/SongX15}, we are using the orientation, center, width, depth and height to encode the 3D bounding box.

\begin{figure}[h]
\begin{center}
   \includegraphics[width=1.0\linewidth]{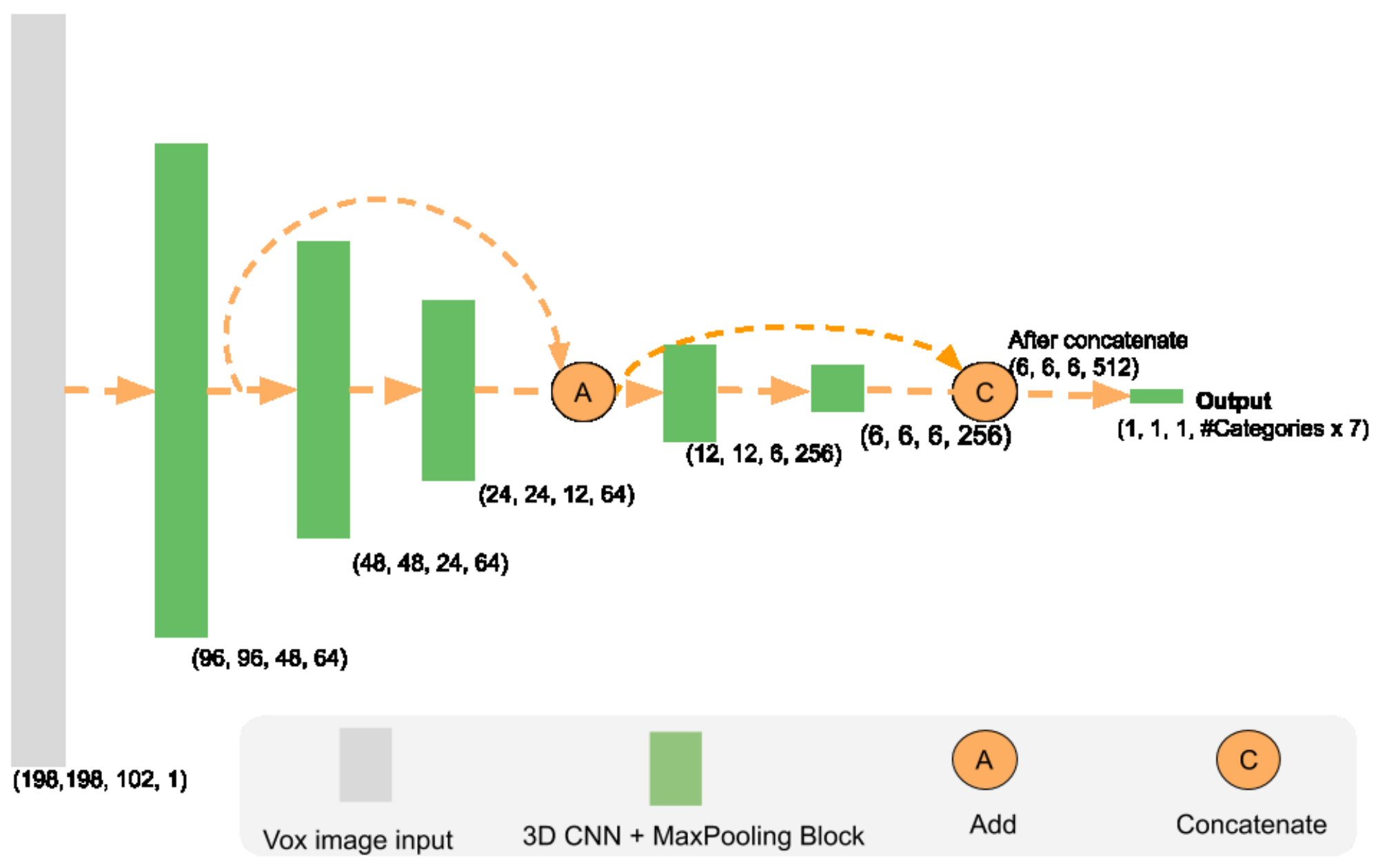}
\end{center}
	\caption{ResnetFCN6 architecture (used for large short scale). Every 3D CNN layer will be followed by a dropout layer. The tensor shape shown here is the output shape of each block. It provides the (width, depth, height, channel) information of the network. The rest three scale networks have the same structure with different input size as shown in Table \ref{big_vox_compare}. The architecture of ResnetFCN35 will be provided in the Supplementary Material. }
        \label{3DCNNnetwork_detection}
\end{figure}

\textbf{Network architecture}
We are using 3D FCN networks to build the 3D detection network by adapting the network structure of ResNet\cite{DBLP:journals/corr/HeZRS15} and Fully Convolutional Network(FCN)\cite{DBLP:journals/corr/LongSD14}. We propose a fast 6 layer fully convolutional 3D CNN model as shown in Figure \ref{3DCNNnetwork_detection}.\\

Inputs of our networks are voxelized images. Our network will have $C*7$ outputs, where $C$ is the number of classes within the corresponding scale network, and 7 is the orientation, center $xyz$ and size(width/depth/height) predictions. The 2D prediction info is implicitly encoded in the system since the prediction is based on each category.\\

\textbf{Loss Function}
We are generating loss function for detection by adjusting the loss function from YOLO9000\cite{DBLP:journals/corr/RedmonF16}. Similar to \cite{DBLP:journals/corr/RedmonF16}, we use simple $L2$ distance instead of Kullback–Leibler divergence to evaluate the difference of predited category probability distributions and the ground truth distributions. For the regression part, for centers, we normalize the $x, y, z$ values to 0 and 1 and then use a sigmoid function to make the prediction. For width($w$), depth($d$) and height($h$), we use anchor to support the prediction. For each category, we set the anchor as the average value of the train2666 samples for objects within this category. The ratio of the bounding box to the related anchors are used to drive the network to make the correct prediction. The formal definition of the loss is given in the formulas below.\\ 
$$L_{detection}^{3D} = \lambda_1 L_{orientation}+\lambda_2 L_{xyz}+\lambda_3 L_{wdh}$$
Where $L_{xyz}=L_x+L_y+L_z, \quad L_{wdh}=L_w+L_d+L_h$, $L_{x}=(x-x^\star)^2$, $L_{y}=(y-y^\star)^2$, $L_{z}=(z-z^\star)^2$, $L_{w}=(\log \frac{w}{a_w}-\log\frac{w^\star}{a_w})^2$, $L_{d}=(\log \frac{d}{a_d}-\log\frac{d^\star}{a_d})^2$, $L_{h}=(\log \frac{h}{a_h}-\log\frac{h^\star}{a_h})^2$. $a_w$, $a_d$, $a_h$ are  width/depth/height of anchors. $\lambda_1$, $\lambda_2$, $\lambda_3$ are used to balance losses.\\ 
\section{Training Process}
For the 2D detection, we are using ResNet\cite{DBLP:journals/corr/HeZRS15} 101 layer as the backbone and using the feature pyramid layers proposed by\cite{DBLP:journals/corr/LinDGHHB16} which is based on Faster RCNN\cite{DBLP:conf/nips/RenHGS15} approach. The loss is the same as\cite{DBLP:journals/corr/LinDGHHB16}. For the 2D detection, the network is pretrained on COCO dataset. Then it is retrained on SUN-RGBD dataset based on RGB or DHS images. Although, the DHS images are different to the RGB images, we find the pretrained weights can still speed up the whole training process and improve the detection results. Data is augmented by adding gaussian blur, random cropping and image translating up to $10\%$ of the original images. 

For the 3D detection, we use the stochastic gradient descent(SGD) with learning rate of 0.01 and a scheduled decay of 0.00001. For regulation we use batch normalization\cite{DBLP:journals/corr/IoffeS15}. The cloud points are randomly rotated around z-axis and jittered during the voxelization process before feeding them to the network.
\section{Efficiency boost by Pipelining}
\begin{figure}[t]
\begin{center}
   \includegraphics[width=1.0\linewidth]{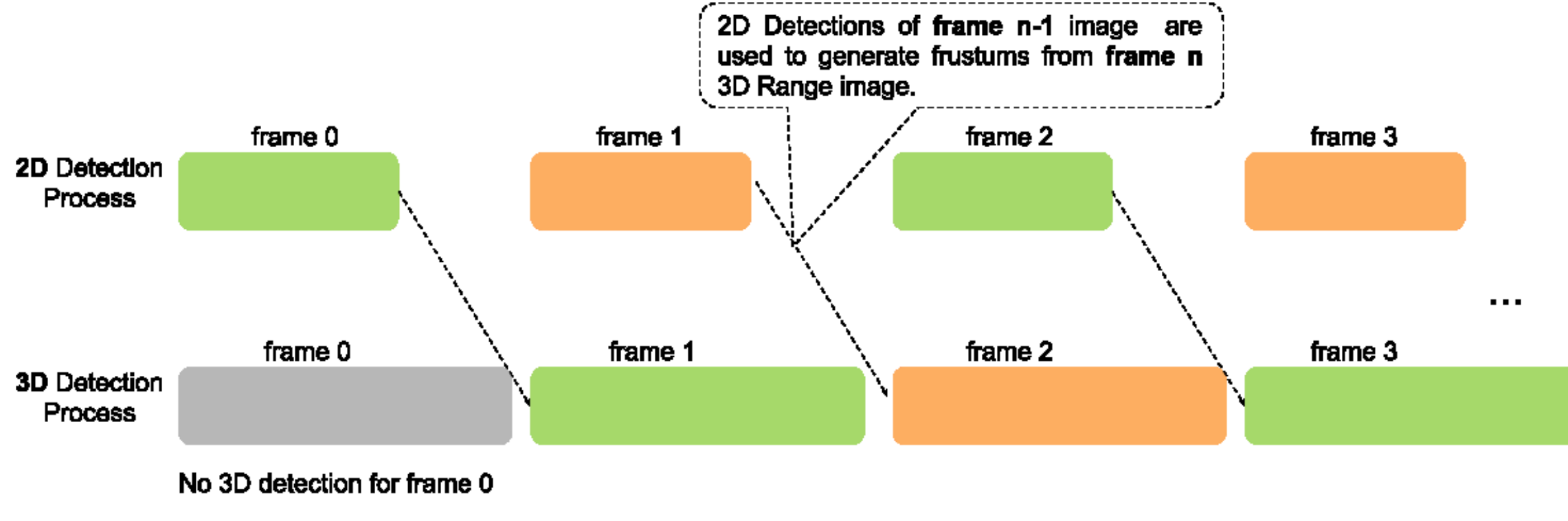}
\end{center}
        \caption{Illustration of using pipelining to speedup the whole detection framework.}
        \label{pipeline}
\end{figure}

 Pipelining instructions is a technology used in central processing units to speed up the computing. An instruction pipeline reads an instruction from the memory while previous instructions are being executed in other steps of the pipeline. Thus multiple instructions can be executed simultaneously. Pipelining can be perfectly used in our system as we have two stages, one is 2D detection and one is 3D detection. In the 3D detection, instead of using the 2D detection of frame n, we can use the 2D detection results of frame n-1 and generate frustums based on that. By using pipelining, our system can be sped up from $t_{2D}+t_{3D}$ to $\max(t_{2D}, t_{3D})$, where $t_{2D}$ and $t_{3D}$ are the 2D and 3D detection time, respectively. The disadvantage of using pipelining is frustums generated from the previous 2D image maybe not accurate under the fast movement of the sensor of an object of interest. However, our system will not suffer significantly as our results show, due to robustness on frustum location. We use multiple candidates with different centers during training to make it robust. Meanwhile, the double frustum method used in our system makes our 3D detections robust to slightly moved 2D detections. The illustration of the pipelining method is shown in figure \ref{pipeline}. By using pipelining, our system can be sped up to 48 ms (this is about $2.5\times$ speedup to the state-of-the-art \cite{DBLP:journals/corr/abs-1711-08488}) when use YOLO v3 and ResNetFCN6. It can achieve 21 frames per second which can well support real time 3D object detection. 

\section{Experiments Results}

\subsection{Effects of Batch Normalization \cite{DBLP:journals/corr/IoffeS15}, Group Normalization \cite{DBLP:journals/corr/abs-1803-08494} and Dropout\cite{JMLR:v15:srivastava14a}}
\begin{figure}[h]
\begin{center}
   \includegraphics[width=1.0\linewidth]{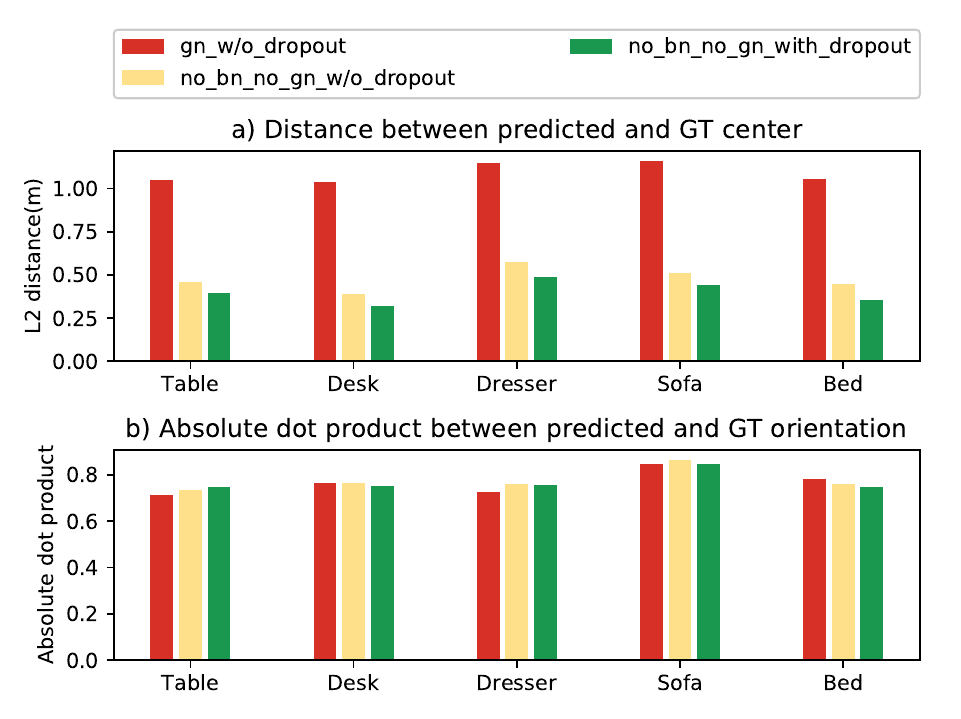}
\end{center}
	\caption{Performance comparison of different combinations on using BN, GN and dropout. ``gn\_w/o\_dropout'' means using GN without dropout.
	``no\_bn\_no\_gn\_w/o\_dropout" means using none. ``no\_bn\_no\_gn\_with\_dropout" means not using BN/GN, however, the Dropout is used.}
        \label{bn_gn_dropout}
\end{figure}

Dropout is a powerful tool to prevent neural networks from overfitting. Batch Normalization(BN) \cite{DBLP:journals/corr/IoffeS15} is another method we can use to speed up the training and prevent overfitting. However, BN performs better when the batch size is large enough.  Since Frustum VoxNet is using 3D CNNs, large batch sizes are not well supported when single GPU is used. Some new technologies are introduced to address the small batch size problem such as Group Normalization(GN) \cite{DBLP:journals/corr/abs-1803-08494}. We explore the performance of different combinations of these methods by evaluating the performance of center and orientation predictions. Results are shown in Figure \ref{bn_gn_dropout}. We do not use BN as our batch size is small and the using of BN will lead to inconsistencies between training and inference. Although when using the GN, there are no inconsistencies between training and inference, the performance of center prediction is worse compared with not using any normalization. Therefore, our final model does not use any normalization. However, dropout is used in our final model as the performance of center prediction is improved.

%

\subsection{Evaluation of the whole system}

\begin{table*}
\begin{center}
	\resizebox{\textwidth}{!}{
\begin{tabular}{|l|c|c|c|c|c|c|c|c|c|c|c|c|c|c|c|c|}
\hline
	& bed     &toilet&  \shortstack{night \\stand}&     bathtub&        chair  &dresser &sofa& table & desk&	bookshelf	&\shortstack{sofa\\ chair}&\shortstack{kitchen\\ counter}&\shortstack{kitchen\\ cabinet}&\shortstack{garbage\\ bin}&microwave&sink\\
\hline\hline
	RGB-D RCNN\cite{DBLP:journals/corr/GuptaGAM14}(RGB-D)&76.0&   69.8&37.1&   49.6&   41.2   &31.3&42.2&43.0& 16.6&   34.9 &N/A&N/A&N/A&46.8&N/A&41.9\\
        2D-driven\cite{Lahoud_2017_ICCV}(RGB)&74.5&  86.2& 49.5&   45.5&   53.0&29.4&49.0&42.3&	22.3&	45.7&	N/A&N/A&N/A&N/A&N/A&N/A\\
	Frustum PointNets\cite{DBLP:journals/corr/abs-1711-08488}(RGB)&56.7&   43.5& 37.2&   81.3&   64.1&33.3&57.4&49.9&	77.8&	67.2&	N/A&N/A&N/A&N/A&N/A&N/A\\
	OURS(RGB)&\textbf{81.0}&  \textbf{89.5}& 35.1&   50.0&   52.4&21.9&53.1&37.7	&18.3	&40.4	&\textbf{47.8}&\textbf{22.0}&\textbf{29.8}&\textbf{52.8}&\textbf{39.7}&31.0\\
	OURS(D)&\textbf{78.7} & 77.6& 34.2&   51.9  &51.8&16.5&48.5	&34.9	&14.2	&19.2 &48.7&19.1&18.5&30.3&22.2&30.1\\
\hline
\end{tabular}}
\end{center}
	\caption{\textbf{2D detection results based on SUN-RGBD validation set}. Evaluation metric is average precision with 2D IoU threshold of 0.5.}
	\label{2d_res}
\end{table*}

\begin{table*}[h]
\begin{center}
        \resizebox{1.0\textwidth}{!}{
\begin{tabular}{|l|cccccccccccc|cc|c|}
\hline
	&  bed     &toilet& \shortstack{night \\stand}&    bathtub&        chair   &dresser&sofa& table & desk&   bookshelf&      \shortstack{sofa\\ chair}&\shortstack{garbage\\ bin}&\shortstack{frustum proposal \\runtime}&\shortstack{3D detection \\runtime}&\shortstack{Total\\ runtime}\\
\hline\hline
        DSS\cite{DBLP:journals/corr/SongX15}(RGB-D)&78.8&   78.9&15.4&   44.2&   61.2    & 6.4   &53.5&50.3& 20.5&   11.9&  N/A&N/A&  N/A&N/A&19.55s\\
	COG\cite{COG}(RGB-D)&63.7&   70.1&27.4&   58.3&   62.2    &15.5&51.0&51.3& 45.2&   31.8&  N/A&N/A&  N/A&N/A&10-30min\\
        2D-driven\cite{Lahoud_2017_ICCV}(RGB-D)& 64.5&   80.4&41.9&   43.5&   48.3    &15.5&50.4&37.0& 27.9&31.4& N/A&N/A& N/A&N/A&4.15s\\
	Frustum PointNets\cite{DBLP:journals/corr/abs-1711-08488}(RGB-D)& 81.1&   90.0&58.1&   43.3&   64.2    &32.0&61.1&51.1& 24.7&   33.3  &N/A&N/A& \textbf{60ms}&60ms&\textbf{0.12s}\\
\hline\hline
	OURS RGB-D (FPN+3D ResNetFCN6)&78.5  &84.5&34.5   &42.4& 47.2 &18.2& 40.3 &30.4 &12.4 &18.0 &47.1&47.6&  110ms&\textbf{48ms}&0.16s\\
	OURS RGB-D (FPN+3D ResNetFCN35)&79.5  &84.6&36.2   &44.6& 49.1 &19.6& 40.8 &27.5 &12.5 &19.1 &47.9&\textbf{48.2}& 110ms&128ms&0.24s\\
	OURS Depth only (FPN+3D ResNetFCN6)& 77.1  &76.1 & 32.4 &42.0 &45.9  & 14.1&35.8&25.3&11.7&16.8 &48.5&35.0& 110ms&\textbf{48ms}&0.16s\\
	OURS Depth only (FPN+3D ResNetFCN35)& 77.4  &76.8 & 33.1 &43.7 &45.8  & 15.2&37.3&25.5&11.8&17.4 &\textbf{48.8}&35.4& 110ms&148ms&0.24s\\
\hline
\end{tabular}}
\end{center}
        \caption{\textbf{3D detection results on SUN-RGBD validation set.} Evaluation metric is average precision with IoU threshold of 0.25 as proposed by\cite{Song_2015_CVPR}. Both COG\cite{COG} and 2D-driven\cite{Lahoud_2017_ICCV} are using room layout context to boost performance while ours, DSS\cite{DBLP:journals/corr/SongX15} and Frustum PointNets\cite{DBLP:journals/corr/abs-1711-08488} are not. Frustum PointNets\cite{DBLP:journals/corr/abs-1711-08488} is using the 3D segmentation information to train the network to boost the 3D detection, while our system and DSS\cite{DBLP:journals/corr/SongX15} are not.}
        \label{3d_res}
\end{table*}

First we evaluate the 2D detector in Table \ref{2d_res}. The evaluation is based on the standard mAP metric with IoU threshold of 0.5. Comparing our RGB-based and depth-based (DHS image) 2D detection, we see that in most cases RGB performs better, but the depth-based 2D detector is competitive. For few classes such as bathtubs, DHS results are slightly better. The reason might be that some classes such as bathtubs have special geometric shapes and they are easier to be detected by depth sensors. Comparing with state-of-the-art methods, our 2D detector performs better in some categories, and we are also introducing new categories. We are on par with most other categories, except for bathtub, desk, and bookshelf.

Full 3D detection results are shown in Table \ref{3d_res}. We provide various variations in our system. First two variations include RGB 2D detector, and the last two include depth only (DHS) 2D detector. In all cases, we use a FPN for the 2D detector. For the 3D detection we have experimented with ResNetFCN6 and ResNetFCN35. As in the 2D case, our 3D detector is on par in most categories with the state-of-the-art, and we have also incorporated more classes. Looking at the computational performance of the 3D detector only, we see that our implementation using ResNetFCN6 provides significant improvements on inference time. Since the architecture is modular (i.e. we can swap out our 2D detector with one from the reported as state-of-the-art), we see that our approach can lead to significant efficiency improvements, without a significant drop in detection accuracy. That will lead to a system geared to real-time robotics applications.

We have also evaluated the efficiency and accuracy of our system when a very fast 2D detector (Yolo v3) is being used. Table \ref{yolo_detection} shows the decrease in detection accuracy as expected. Finally Table \ref{infer_time_comp} provides a detailed analysis of multiple network combinations in terms of efficiency, along with the number of parameters to tune. As mentioned before we can achieve faster inference times in 3D detection, and can thus lead to a faster system overall if we swap our 2D detector with the ones reported as state-of-the-art. Using Yolo and pipelining approach, we can provide a significant boost in total efficiency, with accuracy loss though. 

%
\begin{table}[h]
\begin{center}
        \resizebox{\linewidth}{!}{
\begin{tabular}{|l|cc|ccccc|}
\hline
	&2D network& 3D network & bed     &toilet&   chair   &sofa& table\\
\hline\hline
	\multirow{2}{*}{2D Detection} &FPN&&81.0  &89.5&52.4 &53.1 &37.7 \\
	 &YOLO v3&&71.8  &73.7&38.5 &51.4 &22.1 \\
\hline\hline
	\multirow{2}{*}{3D Detection} &FPN&3D ResNetFCN6&78.5  &84.5&47.2 &40.3 &30.4 \\
	 &YOLO v3&3D ResNetFCN6&66.9  &69.8&30.1 &37.9 &18.8 \\
\hline
\end{tabular}}
\end{center}
        \caption{2D/3D detection results based on YOLO v3 V.S. FPN. 2D detection is based on RGB images. 3D detection is based on RGB-D images.}
        \label{yolo_detection}
\end{table}

\begin{table}[th]
\scriptsize
\begin{center}
               \resizebox{1.0\linewidth}{!}{
\begin{tabular}{|c|c|c|c|c|c|}
\hline
        \multirow{2}{*}{Methods}        &\multicolumn{2}{|c|}{\shortstack{$\#$ parameters}}&\multicolumn{3}{|c|}{\shortstack{Runtime (ms)}}\\
        \hhline{~-----}
	&\shortstack{Frustum\\ proposal}& \shortstack{3D\\ detection}    &\shortstack{Frustum\\ proposal}  &\shortstack{3D \\detection}
	&\shortstack{Total}\\
\hline\hline
      \shortstack{Frustum PointNets (FPN+Pointnet v1)}     &\textbf{28M}&  19M     &60        &60  &120\\
        \shortstack{Frustum PointNets (FPN+Pointnet v2)}     &\textbf{28M}&  22M     &60        &107   &167\\
\hline\hline
	Ours w/o Pipeline (FPN+3D ResNetFCN6)   &42M    &\textbf{2.5M}   &110      &\textbf{48}    &158\\
	Ours w/o Pipeline (FPN+3D ResNetFCN35)   &42M    &23.5M   &110      &149    &259\\
	Ours w/o Pipeline (YOLO v3+3D ResNetFCN6)   &N/A   &\textbf{2.5M}   &\textbf{29}      &\textbf{48}    &\textbf{77}\\
	Ours with Pipeline (YOLO v3+3D ResNetFCN6)   &N/A   &\textbf{2.5M}   &\textbf{29}      &\textbf{48}    &\textbf{48}\\
\hline
\end{tabular}}
\end{center}
\caption {Number of parameters and inference time comparison between Frustum Pointnet and our system. For YOLO v3, input resolution is 416 by 416  and the model FLOPS is 65.86 Bn. }
\label{infer_time_comp}
\end{table}

\section{Conclusion and Future Work}
We presented a 2D-based 3D detection system by using 2D/3D CNNs. Our method can operate in both Depth only and RGB-D sensor modalities. We provide comparable results to state-of-the-art, but with significantly more efficient 3D detection. This is due to the use of networks with fewer number of parameters than competing methods. It is also due to our ability to voxelize only parts of the 3D frustums. This leads to decreased memory requirements and improved resolution around the objects of interest. In future work we will be integrating segmentation that we believe will further boost the detection accuracy of our system.

\section{Acknowledgement}
This work was partially supported by NSF Award CNS-1625843 and Google Faculty Research Award 2017 (special thanks to Aleksey Golovinskiy, Tilman Reinhardt and Steve Hsu for attending to all of our needs). We acknowledge the support of NVIDIA with the donation of the Titan-X GPU used for this work. We thank Jaspal Singh for data preparation and earlier discussion. We also would like to thank Allan Zelener, James Kluz, Jaime Canizales and Bradley Custer for helpful comments and advice. 
\newpage

{\small
\bibliographystyle{ieee}
\bibliography{egbib}
}
\newpage
\section{Supplementary Material}

\subsection{2D detection performance of using RGBDHS as input}
\begin{table*}[th]

\begin{center}
	\resizebox{\textwidth}{!}{
\begin{tabular}{|l|c|c|c|c|c|c|c|c|c|c|c|c|c|c|c|c|}
\hline
	& bed     &toilet&  \shortstack{night \\stand}&     bathtub&        chair  &dresser &sofa& table & desk&	bookshelf	&\shortstack{sofa\\ chair}&\shortstack{kitchen\\ counter}&\shortstack{kitchen\\ cabinet}&\shortstack{garbage\\ bin}&microwave&sink\\
\hline\hline
	OURS(RGB)&\textbf{81.0}&  \textbf{89.5}& 35.1&   50.0&   52.4&21.9&53.1&37.7	&18.3	&40.4	&\textbf{47.8}&\textbf{22.0}&\textbf{29.8}&\textbf{52.8}&\textbf{39.7}&31.0\\
	OURS(D)&\textbf{78.7} & 77.6& 34.2&   51.9  &51.8&16.5&48.5	&34.9	&14.2	&19.2 &48.7&19.1&18.5&30.3&22.2&30.1\\
	OURS(RGB-D)&59.7&  55.0&7.0&    46.5&   37.7&4.7&22.7&25.8	&4.7	&11.1	&32.0&16.2&16.9&28.7&10.2&17.7\\
\hline
\end{tabular}}
\end{center}
	\caption{\textbf{2D detection results based on SUN-RGBD validation set when using different modalities as input}. Evaluation metric is average precision with 2D IoU threshold of 0.5. For the RGB-D one, the model is trained from scratch. The based on RGB and based on D ones are finetuned based on model pretrained on COCO dataset.}
	\label{2d_res_rgbdhs}
\end{table*}
Besides of using the RGB and DHS (from depth sensor) as inputs for the 2D detection framework, we also explored using both the RGB and DHS images to the adjusted model by feeding a six channel (RGBDHS) image to the model. Since the input channel is changed from 3 to 6, we can not use the pretrained weights from the COCO dataset, which cause the results of this RGBDHS one drop a lot. Details can be found in table \ref{2d_res_rgbdhs}. We also tried only random initialize the first layer of the model and kept the rest weights unchanged, however, this cause issues during the training process. As next steps, we can try copy the RGB weights to the DHS related weights at the initial layers to see the performance.\\

\subsection{Visualizations of 2D and 3D detection results}
\begin{figure*}[p]
\begin{center}
        \includegraphics[width=1.0\textwidth]{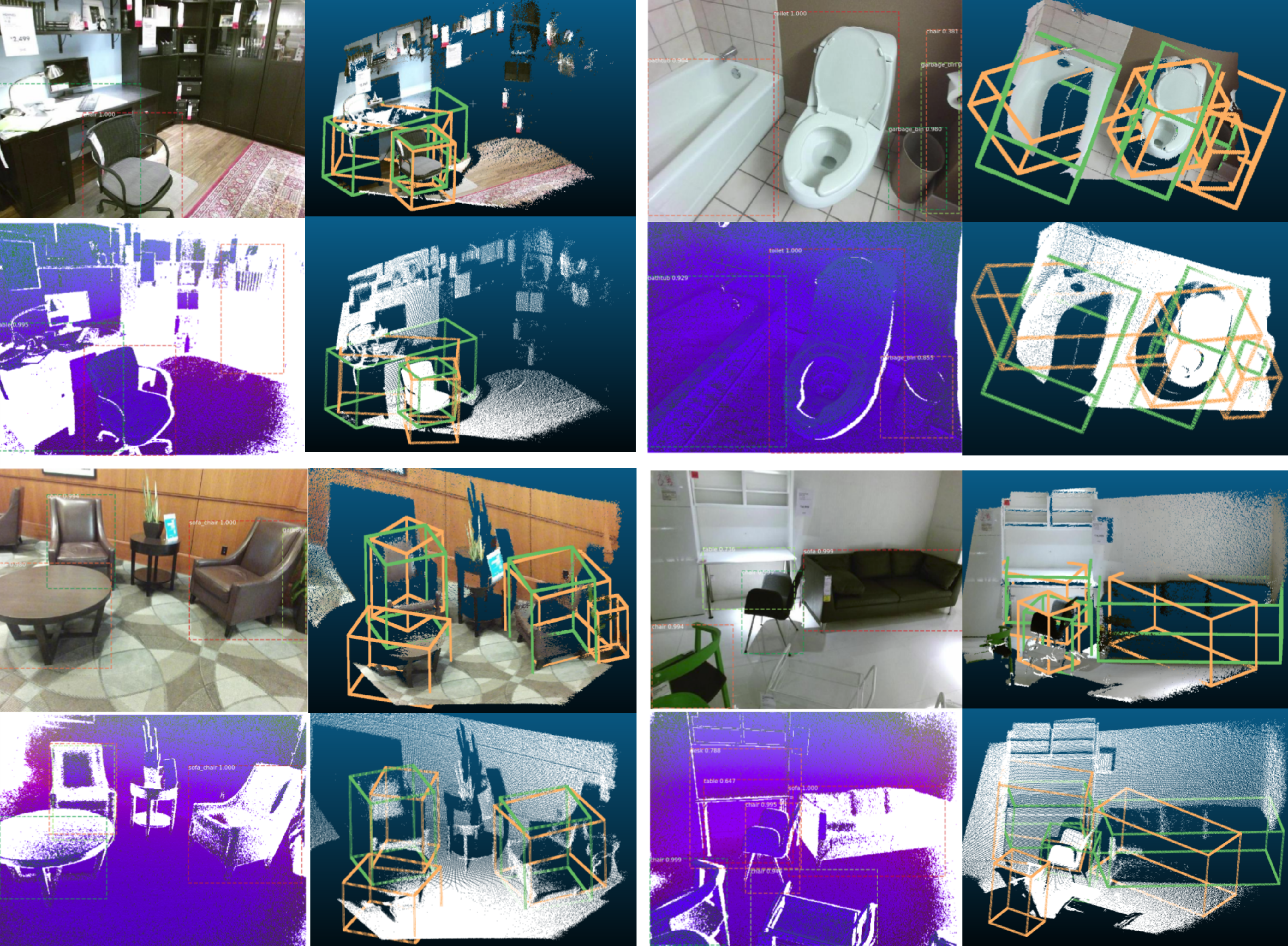}
\end{center}
	\caption{Visualizations of 2D and 3D detection results part 1. This visualization contains four images. For each image, upper left shows 2D detection based on RGB image.  Upper right shows the corresponding 3D detection results (light green ones are the 3D ground truth boxes and orange-colored boxes are predictions) based on frustums generated from RGB image 2D detections (to have a better visualization, RGB colors are projected back to the cloud points). Lower left shows 2D detection based on DHS image. Lower right shows the corresponding 3D detection results (light green ones are the 3D ground truth boxes and orange-colored boxes are predictions) based on frustums generated from DHS image 2D detections. For the first image in the first row, our system can perfectly detect the chair. For the desk, the orientation is off as the frustum generated by the 2D bounding box contains some cloud points from the chair. For the second image, we can see that the based on RGB image system detect more false positive objects in the 2D stage and hence more 3D false positive objects will be detected. For the first image of the second row, our system successfully detect the unlabelled table. For the last image, the sofa's orientation is off as there are too many points are missing for the sofa.}
        \label{f1}
\end{figure*}

\begin{figure*}[p]
\begin{center}
        \includegraphics[width=1.0\textwidth]{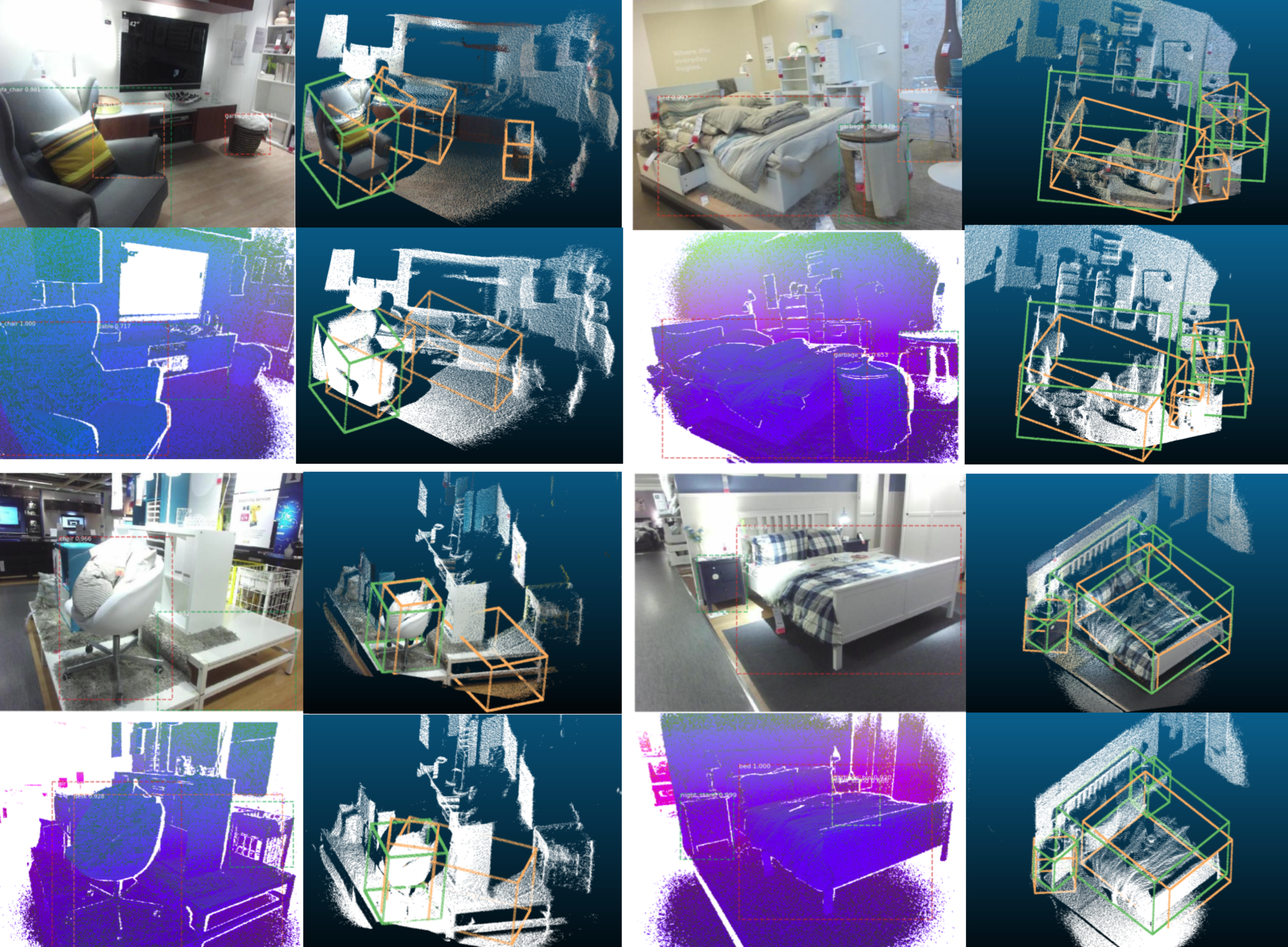}
\end{center}
	\caption{Visualizations of 2D and 3D detection results part 2. This visualization contains four images. Please read the caption of Figure \ref{f1} to get an explanation about how to understand the visualization. On the left part, our system can successfully detect unlabelled object such as garbage bin and table. On the top right image, our system fails to detect one table in 2D detection stage as it is partially observed. For the last one, one night stand is undetected as it is blocked by bed.}
        \label{f2}
\end{figure*}

Visualizations of both 2D and 3D detection results are shown in Figure \ref{f1} and Figure \ref{f2} for both the based on RGB-D system and based on depth image only system. From Figure \ref{f1} we can see that the 3D detection system works well for both the based on RGB-D and based on Depth only systems. The RGB-D based 3D detection system will generate some false positive 3D detections as it has more false positive detection during 2D detection stage. We can also find out that our system can detect objects which were not labeled during the data annotation. In Figure \ref{f2}, in the left two images, our system can successfully detects unlabelled objects. On the right two images of Figure \ref{f2}, we can see we have some false negative detections as there are too few points within the object. Detail explanations are given in captions.\
\

\subsection{Proof of equation 1}
\begin{proof}
Define the threshold used for positive 3DCB as $threshold_{3D}$, and a 3DCB is positive when $IoI^{3D} = IoI^{XY} * IoI^{Z} \geq threshold_{3D}$. The $recall^{volume}$ $recall^{XY}$, $recall^{Z}$, $threshold_{XY}$ and  $threshold_{Z}$ are defined in the main article. We set the $ threshold_{3D}=threshold_{XY}*threshold_{Z}$.\\
	As $3DCB^{positive}_{XY} \cap 3DCB^{positive}_{Z}$ implies $IoI^{XY}\geq threshold_{XY}$ and $IoI^{Z} \geq threshold_{Z}$. We can further get $IoI^{3D} = IoI^{XY} * IoI^{Z} \geq threshold_{XY}*threshold_{Z}=threshold_{3D}$, which implies 3DCBs in the set of $3DCB^{positive}_{XY} \cap 3DCB^{positive}_{Z}$ are positive. Meanwhile, we can show from an example that the set of $3DCB^{positive}$ can possibly be obtained from $3DCB^{positive}_{XY} \cap 3DCB^{non positive}_{Z}$, where $3DCB^{non positive}_{Z}$ is a complement set of $3DCB^{positive}_{Z}$: if $threshold_{XY}=0.9,\quad threshold_{Z}=2.9$, we can get $threshold_{3D}=0.81$. A 3DCB with $IoI_{XY}=1.0,\quad IoI_{Z}=0.82$ will be an element of set $3DCB^{positive}_{XY} \cap 3DCB^{non positive}_{Z}$. Also it is a positive 3DCB. From above arguments, we can conclude the following relation:\\
\begin{equation}  \label{eq:2}
 3DCB^{positive} \supseteq \{3DCB^{positive}_{XY} \cap 3DCB^{positive}_{Z}\}
\end{equation}
From equation \ref{eq:2}, we can get:\\

\begin{equation}  \label{eq:3}
|3DCB^{positive}| \geq |3DCB^{positive}_{XY} \cap 3DCB^{positive}_{Z}|
\end{equation}
We can also rewrite right part of equation \ref{eq:2} as:\\
\begin{equation}  \label{eq:4}
	\begin{split}
		\{3DCB^{positive}_{XY} &\cap 3DCB^{positive}_{Z}\} \\
		&= 3DCB^{positive}_{XY} \setminus \\
		&\quad \{3DCB^{positive}_{XY} \cap \{3DCB \setminus 3DCB^{positive}_{Z} \} \} 
	\end{split}
\end{equation}
From equation \ref{eq:4}, we can further get:\\
\begin{equation}  \label{eq:5}
        \begin{split}
		&|3DCB^{positive}_{XY} \setminus \\
		&\quad \{3DCB^{positive}_{XY} \cap \{3DCB \setminus 3DCB^{positive}_{Z} \} \}|\\
		&\quad \geq |\{3DCB^{positive}_{XY} \setminus  \{3DCB \setminus 3DCB^{positive}_{Z} \}|\\
		&\quad \geq |3DCB^{positive}_{XY}|-|3DCB \setminus 3DCB^{positive}_{Z}|\\
		&\quad = |3DCB^{positive}_{XY}|-(|3DCB|- |3DCB^{positive}_{Z}|)
        \end{split}
\end{equation}
From equations \ref{eq:3}, \ref{eq:5}, we can get:\\ 
\begin{equation}  \label{eq:6}
        \begin{split}
		|3DCB^{positive}| &\geq |3DCB^{positive}_{XY}|-(|3DCB|-|3DCB^{positive}_{Z}|)\\
		&=|3DCB^{positive}_{XY}|+|3DCB^{positive}_{Z}|-|3DCB|
        \end{split}
\end{equation}
From equation \ref{eq:6}, we can get:\\
\begin{equation}  \label{eq:7}
	      \begin{split}
		      &\frac{|3DCB^{positive}|}{|3DCB|}\\
		      &\quad \geq \frac{|3DCB^{positive}_{XY}|+|3DCB^{positive}_{Z}|-|3DCB|}{|3DCB|}\\
	      \end{split}
	\end{equation}
Equation \ref{eq:7} can be rewritten as:\\
\begin{equation}  \label{eq:8}
 recall^{volume} \geq  recall^{XY} +recall^{Z} -1
\end{equation}
Since $recall^{volume}$ is supposed to be greater or equal to 0, we get:\\
\begin{equation}  \label{eq:9}
 recall^{volume} \geq \max(0, \quad recall^{XY} +recall^{Z} -1)
\end{equation}
\end{proof}
\subsection{Evaluate Frustum VoxNet Results based on Ground Truth 2D Bounding Box}
\begin{figure*}[p]
\begin{center}
   \includegraphics[width=1.0\textwidth]{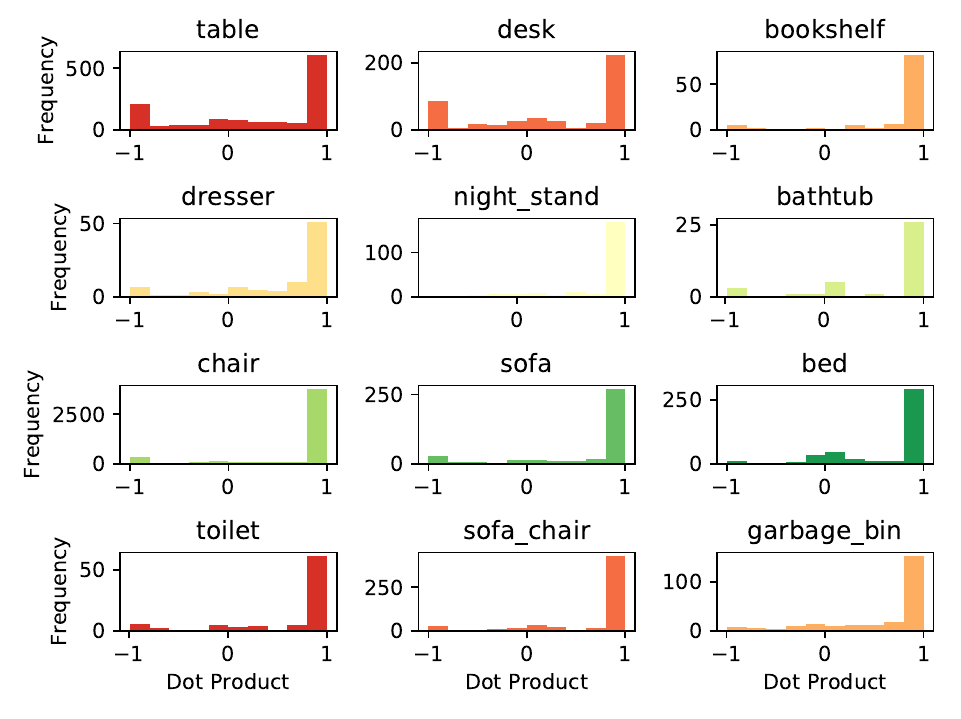}
\end{center}
        \caption{Histogram of the dot product between predicted orientation and GT orientation. Histograms are not normalized.}
        \label{orientation_hist_more}
\end{figure*}

\begin{figure*}[p]
\begin{center}
   \includegraphics[width=1.0\textwidth]{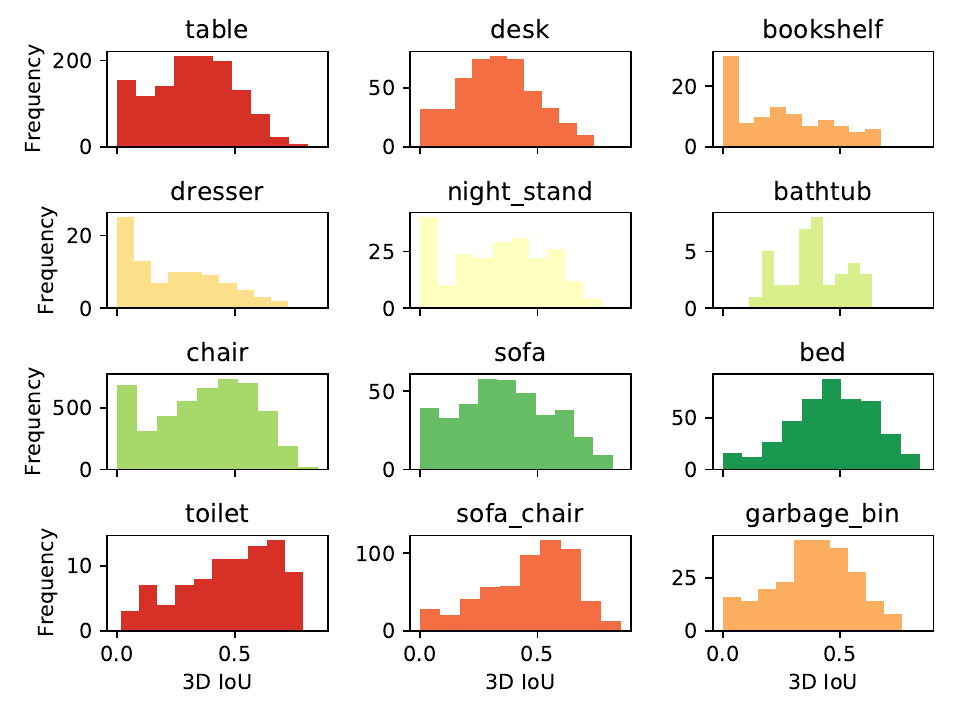}
\end{center}
        \caption{Histogram of 3D IoU. Histograms are not normalized.}
        \label{iou_hist}
\end{figure*}
\subsubsection{Evaluation Metrics}
We are using following metrics to evaluate the predictions based on ground truth 2D bounding boxes.
$$D_x=|x^*-x|, \quad D_y=|y^*-y|, \quad D_z=|z^*-z|$$
$$D_w=|w^*-w|, \quad D_d=|d^*-d|, \quad D_h=|h^*-h|$$
$$D_{xyz}=\sqrt{(x^*-x)^2+(y^*-y)^*+(z^*-z)}|$$
$$D_{wdh}=\sqrt{(w^*-w)^2+(d^*-d)^*+(h^*-h)}|$$
$x^*, y^* , z^*$ are the predicted center and $x, y, z$ are ground truth.  $w^*$, $d^*$, $h^*$ are the predicted width/depth/height and $w, d, h$ are ground truth.\\
\subsubsection{Evaluation Results}
\begin{table}[p]
\scriptsize
\begin{center}
                               \resizebox{1.0\linewidth}{!}{
\begin{tabular}{|c|c|c|c|c|c|}
\hline
	& &$\overline{x-x^*}$&$\overline{y-y^*}$ &$\overline{z-z^*}$ &$\overline{D_{xyz}}$\\
\hline\hline
\multirow{2}{*}{Table}	&Frustum Average Center&-0.005&	-0.233&	0.075&	0.522\\
	&Predicted from Frustum VoxNet&0.014&	-0.040&	0.030&	\textbf{0.395}\\
\hline
\multirow{2}{*}{Desk}	&Frustum Average Center&-0.010&	-0.198&	0.109&	0.428\\
	&Predicted from Frustum VoxNet&0.028&	-0.040&	0.048&	\textbf{0.319}\\
\hline
\multirow{2}{*}{Sofa}	&Frustum Average Center&-0.015&	-0.168&	0.010&	0.516\\
	&Predicted from Frustum VoxNet&0.007&	0.041&	0.013&\textbf{	0.444}\\
\hline
\multirow{2}{*}{Bed}	&Frustum Average Center&0.031&	-0.195&	0.013&	0.573\\
	&Predicted from Frustum VoxNet&-0.009&	0.010&	-0.012&	\textbf{0.354}\\
\hline
\end{tabular}}
\end{center}
\caption {Result comparison predicted between frustum center and predicted center from Frustum VoxNet.}
\label{average_xyz}
\end{table}
We compare the center prediction based on the frustum average center and the prediction from our Frustum VoxNet system. Table \ref{average_xyz} provides the average distance between predicted and ground truth centers by using these two methods. As expected, the Frustum VoxNet prediction is better than the average center from frustum. 
Evaluation results for the performance of Frustum Voxnet based on frustums generated from ground truth bounding boxes are shown in Table \ref{detail_eva_based_on_gt_more}. Histograms of the dot product between predicted orientations and GT orientations for each category are shown in Figure \ref{orientation_hist_more}. Histograms of 3D detection IoU for each category are shown in Figure \ref{iou_hist}. \\
\begin{table*}[t]
\begin{center}
        \resizebox{\textwidth}{!}{
\begin{tabular}{|c|c|cccc|cccc|c|cc|}
\hline
        category    &\shortstack{instance \\ number}&$\overline{D_x}$&$\overline{D_y}$& $\overline{D_z}$& $\overline{D_{xyz}}$&   $\overline{D_w}$&      $\overline{D_d}$&        $\overline{D_h}$&     $\overline{D_{wdh}}$& $\overline{|o^* \cdot o|}$  &\shortstack{average \\3D IoU} &\shortstack{3D recall\\(IoU@0.25)}\\
\hline\hline
        table&  1269&   0.201&  0.280&  0.070&  0.395&  0.206&   0.132&   0.042&   0.287&       0.747&  0.319&  0.656\\
        desk    &457&   0.158&  0.220&  0.080&  0.319&  0.180&   0.122&   0.052&   0.258&       0.752&  0.329&  0.674\\
        dresser&91&     0.248&  0.298&  0.135&  0.489&  0.126&   0.064&   0.107&   0.209&       0.758&  0.241&  0.451\\
        sofa&   381&    0.213&  0.320&  0.075&  0.444&  0.210&   0.099&   0.048&   0.264&       0.847&  0.459&  0.796\\
        bed&    441&    0.195&  0.220&  0.096&  0.354&  0.154&   0.125&   0.083&   0.246&       0.746&  0.462&  0.898\\

\hline\hline
        night stand&    220&    0.156&  0.226&  0.069&  0.314&  0.050&   0.037&   0.044&   0.087&  0.830&  0.329&  0.655\\
        bathtub&        37&     0.162&  0.114&  0.067&  0.226&  0.134&   0.071&   0.040&   0.173&       0.805&  0.383&  0.811\\
        chair&          4777&   0.118&  0.217&  0.067&  0.286&  0.038&   0.048&   0.047&   0.089&       0.886&  0.369&  0.708\\
        sofa chair&     575&    0.109&  0.168&  0.070&  0.242&  0.058&   0.051&   0.045&   0.103&       0.840&  0.466&  0.849\\
        garbage bin&    248&    0.065&  0.098&  0.050&  0.145&  0.043&   0.035&   0.042&   0.082&       0.760&  0.384&  0.782\\
\hline\hline
        toilet& 87&     0.051&  0.093&  0.073&  0.148&0.028&    0.039&  0.047&  0.076&  0.825&  0.498&  0.929\\
\hline\hline
        bookshelf&      106&    0.183&  0.303&  0.130&  0.433&  0.410&  0.063&  0.149&  0.474&  0.880&  0.345&  0.679\\
        \hline
\end{tabular}}
\end{center}
        \caption{\textbf{Detail evaluation results.} Frustum VoxNet is evaluated based on SUN-RGBD validation set. Frustums used to finalize detection are generated from ground truth 2D bounding boxes. The 3D IoU threshold used for 3D recall is 0.25.}
        \label{detail_eva_based_on_gt_more}
\end{table*}
\subsection{ResNetFCN35 network structure}
\begin{figure*}[p]
\begin{center}
   \includegraphics[width=1.0\textwidth]{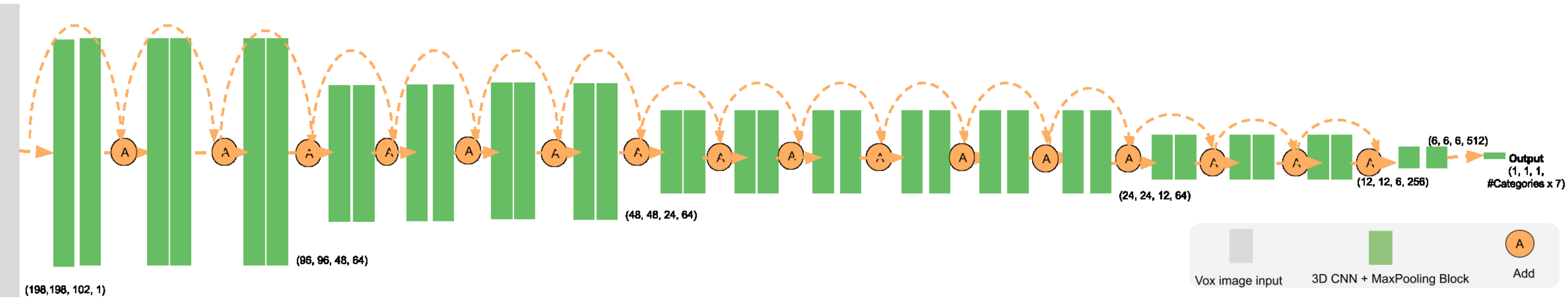}
\end{center}
        \caption{ResNetFCN35 network structure.}
        \label{resnet35}
\end{figure*}
ResNetFCN35 network structure is shown in Figure \ref{resnet35}.\\

\end{document}